\documentclass{llncs}

\usepackage[T1]{fontenc}
\usepackage{graphicx}
\usepackage{hyperref}
\usepackage{xcolor}
\def\BibTeX{{\rm B\kern-.05em{\sc i\kern-.025em b}\kern-.08em 
T\kern-.1667em\lower.7ex\hbox{E}\kern-.125emX}}
\usepackage{multirow}
\usepackage{url}
\usepackage{listings}

\definecolor{lstgreen}{rgb}{0,0.5,0}

\usepackage{subcaption} % Para utilizar subfigure
\begin{document}
\title{Temporal fine-tuning for early risk detection\thanks{Published in the Proceedings of the 53rd JAIIO – ASAID (Argentine Symposium on Artificial Intelligence and Data Science), Bahia Blanca, Buenos Aires, Argentina, 2024. ISSN: 2451-7496, p. 137}}
\author{Horacio Thompson\inst{1,2} \and
Esaú Villatoro-Tello\inst{3} \and
Manuel Montes-y-Gómez\inst{4} \and
Marcelo Errecalde\inst{1}}
\authorrunning{H. Thompson et al.}
\institute{Universidad Nacional de San Luis, San Luis, Argentina \and
Consejo Nacional de Investigaciones Científicas y Técnicas (CONICET) \and
% \email{hjthompson@unsl.edu.ar} \email{merreca@unsl.edu.ar} 
% \email{hjthompson@unsl.edu.ar} - \email{merreca@unsl.edu.ar} \\ \and
Idiap Research Institute, Martigny, Switzerland \and 
Instituto Nacional de Astrofísica, Óptica y Electrónica (INAOE), Puebla, Mexico \\
\smallskip 
\email{hjthompson@unsl.edu.ar}, \email{esau.villatoro@idiap.ch}, \email{mmontesg@inaoep.mx}, \email{merreca@unsl.edu.ar} \\
}
\maketitle              % typeset the header of the contribution
\begin{abstract}
Early Risk Detection (ERD) on the Web aims to identify promptly users facing social and health issues. Users are analyzed post-by-post, and it is necessary to guarantee correct and quick answers, which is particularly challenging in critical scenarios. ERD involves optimizing classification precision and minimizing detection delay. Standard classification metrics may not suffice, resorting to specific metrics such as ERDE$\theta$ that explicitly consider precision and delay. The current research focuses on applying a multi-objective approach, prioritizing classification performance and establishing a separate criterion for decision time. In this work, we propose a completely different strategy, \emph{temporal fine-tuning}, which allows tuning transformer-based models by explicitly incorporating time within the learning process. Our method allows us to analyze complete user post histories, tune models considering different contexts, and evaluate training performance using temporal metrics. We evaluated our proposal in the depression and eating disorders tasks for the Spanish language, achieving competitive results compared to the best models of MentalRiskES 2023. We found that temporal fine-tuning optimized decisions considering context and time progress. In this way, by properly taking advantage of the power of transformers, it is possible to address ERD by combining precision and speed as a single objective.


\end{abstract}

% \newpage
\section{Introduction}
One of the problems that has become relevant in recent years is Early Risk Detection (ERD) on the Web, which consists of correctly identifying risk users as soon as possible. It incorporates a significant complexity to standard classification problems since the users are analyzed post-by-post rather than processing the complete history. 
Challenges such as CLEF eRisk \cite{losada2016,losada2017,losada2018,losada2019,losada2020,parapar2021,parapar2022,Parapar2023} and MentalRiskES \cite{marmol2023} have emerged to solve ERD problems in scenarios where users suffer from different mental disorders. In particular, MentalRiskES 2023 was the first edition to propose tasks exclusively for the Spanish language. 
These challenges defined a testing environment for ERD problems within a post-round scheme. In every round, a post from each user is received, and it is necessary to respond to each to continue to the next round. Whether the user is detected as positive, a risk alarm is issued, and the analysis is ended in that post-round. Otherwise, if the user's responses were successively negative and the post history ends, it is concluded that the user is negative.
The solutions are evaluated considering standard classification metrics, as well as temporal metrics such as ERDE$\theta$ \cite{losada2016} and F-latency \cite{sadeque2018}.

Although precision is essential for ERD problems, as time progresses and decisions are delayed, speed becomes increasingly important, taking priority over precision. Because both are fundamental objectives to solve these problems, two possible approaches arise: multi-objective and combined single-objective. In the first, each objective is solved independently according to the priorities of the problem, typically addressing the precision and then optimizing the decision time. With the emergence of \emph{transformers} \cite{Vaswani2017}, numerous studies have focused on this approach by prioritizing the correct classification of users. 
These models are trained and validated differently than in the testing environment, and this disparity complicates the optimal model selection. 
Moreover, the limited number of tokens in each architecture frequently leads to a partial observability scenario regarding the complete user's post history. 

On the contrary, in a combined single-objective approach, precision and speed are concurrently considered during the learning process. That aspect has significant advantages over the standard (separated) multi-objective approach. First, considering input samples and time progress would allow for defining a single learning component where models can receive and integrate all the information needed for ERD. Furthermore, it would allow a scenario similar to the one used in the testing stage to be elegantly reproduced during training. 
As a consequence, optimal models can be obtained using the same temporal evaluation metrics utilized for assessing the ERD systems. 
However, despite these advantages, few works have applied this approach.

In this work, we propose \emph{temporal fine-tuning}, a novel method that allows tuning transformer-based models and simultaneously optimizing precision and speed for ERD problems. 
The main contributions are: 1) transforming the input from the users to keep track of the delays, 2) defining a loss function to be optimized according to a temporal evaluation metric, and 3) implementing a training and validation procedure that allows selecting the optimal hyperparameters for a particular ERD problem. 
This work presents an innovative strategy that addresses a gap in the research domain, showing promising results that support the viability of our proposal when compared to state-of-the-art methods. In this way, our study offers new perspectives for future research.

\section{Related work}
Several studies have addressed ERD through transformer-based methods that follow the multi-objective approach. Most strategies solve ERD as a user classification problem by applying the fine-tuning process on models such as BERT and RoBERTa \cite{bucur2021early,pan2023umuteam,devaguptam2022early,ramos2023i2c}, as well as pre-trained models for Spanish such as BETO and RoBERTuito \cite{gonzalez2023sinai,pan2023umuteammental,garcia2023improving-transformers}. Other proposals include embeddings extracted from transformers followed by different types of classifiers \cite{marmol2022,rujas2023development,Talha2023,Stalder2022,wu2022}. Although these strategies focused on optimizing classification performance, they also achieved temporal efficiency in an ERD environment, suggesting that classification robustness could impact decision speed. A more representative method within this approach was proposed in \cite{loyola2022,thompson2023strategies,thompson2023early}, addressing precision and speed separately. A BERT-based classifier was applied, followed by an independent component for decision-making, which was not considered within the learning process of the predictor. 
Sun et al. \cite{sun2023ranking} argue that an early classifier should be accurate, but when decisions are unclear and delayed, adding a lookahead component to the classifier is necessary. At this point, the combined single-objective approach gains importance by considering precision and speed concurrently during learning. However, there is a limited number of works that use this approach. A study proposed a memorization network for affective states that is updated considering time \cite{kang2022tua1}, and the EARLIEST architecture for time series was adapted to balance precision and speed in decisions using reinforcement learning \cite{Loyola2021,loyola2022}. In this work, we propose an original strategy that contributes to the research domain by applying the combined single-objective approach to solve ERD problems.

\section{Temporal fine-tuning}
\begin{figure}[!ht]
    % \centering
\includegraphics[width=0.9\textwidth]{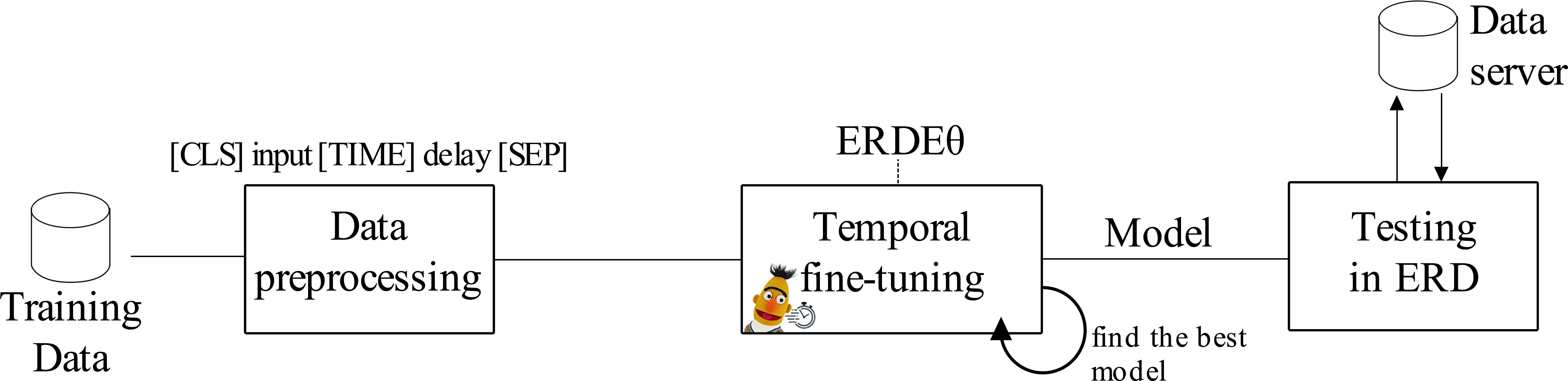} 
\caption{Pipeline of the \emph{temporal fine-tuning} process. The samples are modified including time, \emph{temporal fine-tuning} is applied according to an ERDE$\theta$, and the best model is chosen to be evaluated in an ERD environment.}
\label{fig:pipeline}
\end{figure}

In this section, we will address the most important aspects of our proposal to incorporate time during the learning process and then evaluate the models in an ERD environment. Fig. \ref{fig:pipeline} shows the pipeline of our proposal.

\subsection{Data preprocessing}
Time is explicitly included in the input samples, considering the number of posts that have been read until the moment ($delay$). For an arbitrary sample $\langle$input, label$\rangle$, the new input is defined as input$_{delay}$ = [CLS] input [TIME] $delay$ [SEP]. The [TIME] token is added to the model architecture, separating the post content of the moment it was read. For example, \emph{hoy es un día triste} (today is a sad day) evaluated at $delay = 10$, the model would receive [CLS] \emph{hoy es un día triste} [TIME] $10$ [SEP].

\subsection{\emph{Temporal fine-tuning} process}
Unlike standard classification tasks, where systems are evaluated with metrics such as F1 score, in ERD  problems, the performance of the models is assessed according to specialized metrics that explicitly consider classification accuracy and delay in detecting positive cases. One of the most popular is ERDE$\theta$, defined~as:

\begin{equation}
\mbox{ERDE}_\theta(d,k) = 
\left\{ \begin{array}{ll}
c_{fp} \\
c_{fn}  \\ 
lc_\theta(k) \cdot c_{tp} & \\ 
0 &       
\end{array} \right.
\label{eq:erde}
\end{equation}
\[
\begin{array}{lll}
c_{fp} & : & \mbox{for false positives (FPs)} \\
c_{fn} & : & \mbox{for false negatives (FNs)} \\
lc_\theta(k) \cdot c_{tp} & : & \mbox{for true positives (TPs)} \\
0 & : & \mbox{for true negatives (TNs)}
\end{array}
\]

\medskip \noindent
where the latency cost ($lc_\theta$) is:

\begin{equation} 
lc_\theta(k) =  1 - \frac{1}{1+e^{k-\theta}} 
\label{eq:lc} 
\end{equation}

\medskip \noindent
It means that, for a decision $d$ that is a TP at time $k > \theta$, the penalty will depend on the $c_{tp}$ value, where typically  $c_{tp} = c_{fn} = 1$, i.e., the maximum penalty. In this way, the ERDE$\theta$ metric could be helpful during the learning process since it evaluates the correctness and delay of the final decisions. 

\begin{figure}[htbp]
    \includegraphics[width=0.95\textwidth]{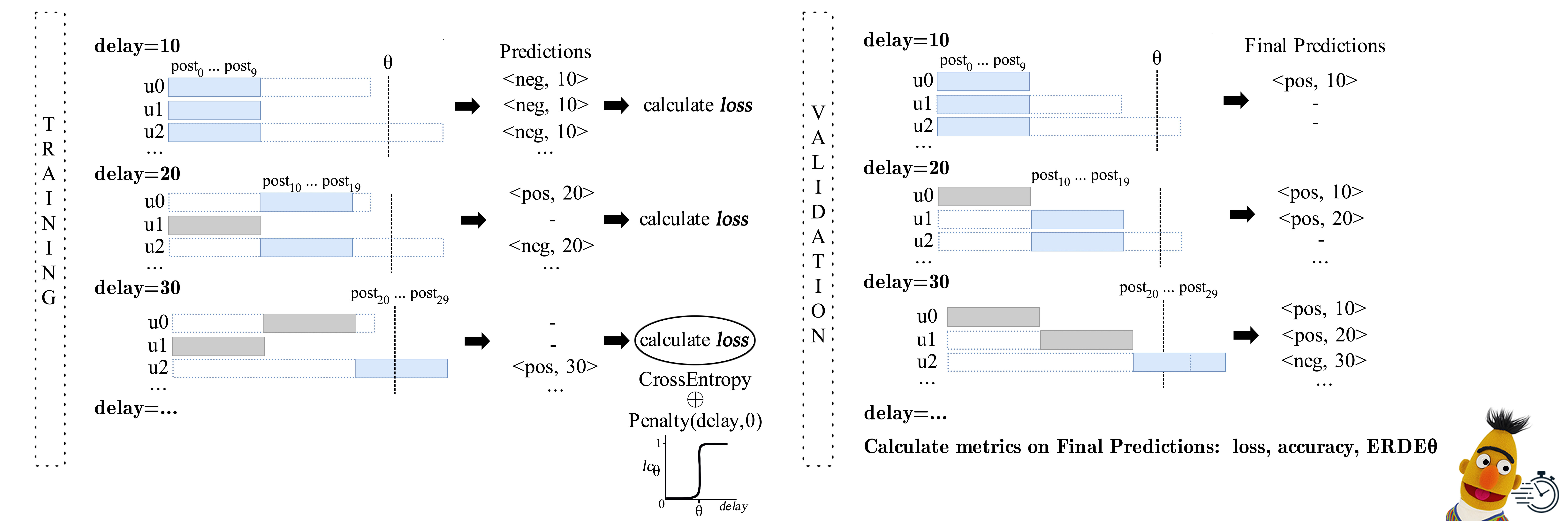}
    \caption{\emph{Temporal fine-tuning} scheme for an epoch. The training and validation stages are subdivided into \emph{delays}, where users are evaluated according to post windows. In training, \emph{loss} is calculated at the end of each \emph{delay} considering CrossEntropy and $lc_\theta$. In validation, the model performance is evaluated considering \emph{loss}, \emph{accuracy}, and ERDE$\theta$.}
    \label{fig:t_ft}
\end{figure}

The \emph{temporal fine-tuning} process is carried out in epochs with the typical training and validation stages. We propose to incorporate in both stages an evaluation scheme similar to the testing environment for ERD problems, where time is measured according to the number of posts requested to issue a response (\emph{delay}). In each \emph{delay}, post windows of length M are evaluated, which consist of concatenating the current post and the previous M-1. The \emph{delays} are configured according to the window size. Fig. \ref{fig:t_ft} shows that, for M=10, \emph{delay}=10 evaluates from post 0 to post 9, \emph{delay}=20 from post 10 to post 19, and continues until all users $u_i$ are analyzed.

The \emph{loss} function is calculated at the end of each \emph{delay}, evaluating according to the post window over those users who have not yet completed their analysis, and the gradient is propagated to the entire transformer.
We used the ERDE$\theta$ metric to design the \emph{loss} function, but it was not included as indicated in (\ref{eq:erde}) since it would not be differentiable. Instead, a linear and differentiable function was implemented by considering the classification performance (CrossEntropy) and harshly penalizing TPs that are delayed (using $\theta$ as a limit, according to (\ref{eq:lc})). In this way, as the \emph{loss} is minimized, ERDE$\theta$ is also reduced, establishing it as the training objective. Listing \ref{lst:loss} shows the procedural details for the \emph{loss} function.
The validation stage is evaluated with the same \emph{delay} scheme, concluding the epoch by calculating the \emph{loss}, \emph{accuracy}, and ERDE$\theta$ metrics, which can be weighted to select the optimal model for an ERD problem.

\begin{lstlisting}[
    language=Python, 
    caption=Implementation of the \emph{loss} function (pseudo-code), 
    captionpos=b,
    label={lst:loss},
    basicstyle=\footnotesize, % \small \footnotesize
    belowskip=-10pt, % Ajusta este valor según sea necesario
    float=ht 
    ]
def temporal_loss(preds_labels, preds_times, 
                    real_labels, real_times, $\theta$):
""" Calculates the temporal loss function.
:param preds_labels: List of 1's and 0's (predicted labels).
:param preds_times: List of integers (#posts read).
:param real_labels: List of 1's and 0's (real labels).
:param real_times: List of integers. (#total posts).
:param $\color{lstgreen}\theta$: Integer. Time limit for optimization (ERDE$\color{lstgreen}\theta$).
:return: Mean of the total loss. """
  # Calculate classification loss
  cls_loss = CrossEntropy(preds_labels, real_labels)
  # Calculate penalty for delay
  delay_loss = []
  for pred, real, pred_time, real_time 
    in zip(preds_labels, real_labels, preds_times, real_times):
    if (pred==1 and pred==real)  # TPs
        and (real_time<pred_time or $\theta$<preds_time): # Delayed
      delay_loss.append(1)
    else
      delay_loss.append(0)
  # Calculate total loss using classification and delay loss
  total_loss = []
  for cls, delay in zip(cls_loss, delay_loss):
    if delay==1:
      total_loss.append(delay)
    else 
      total_loss.append(cls)

  return total_loss.mean()
\end{lstlisting}

\subsection{Model testing in ERD}
The best model obtained can be evaluated using a mock-server tool \cite{loyola2022}, which simulates ERD environments through rounds of posts and answers submissions, and it calculates the final results using different metrics. A client application was defined to interact with the server: when it receives a round of posts, the system preprocesses them by adding time, invokes the predictive model, and returns a response. We used a sliding post window, configured as in the learning stage, and a simple decision policy: if the probability exceeds a \emph{threshold}, a user at-risk alarm is issued; otherwise, the analysis should continue.

\section{Experimental results}
% The experiments were carried out on the \emph{depression} and \emph{eating disorders} tasks of MentalRiskES 2023, using the following datasets (The MentalRiskES Organizers supplied data and cannot be disclosed or shared publicly. More details in \cite{marmol2023}): 
The experiments were carried out on the \emph{depression} and \emph{eating disorders} tasks of MentalRiskES 2023, using the following datasets (supplied by the MentalRiskES Organizers and cannot be disclosed or shared publicly; more details in \cite{marmol2023}): 
\emph{train} and \emph{trial} to train the models and \emph{test} to evaluate them with the mock-server and compare results with other teams (Table \ref{fig:datasets}). It can be noted that they present a relatively balanced distribution between classes. In addition, a model with an acceptable performance should complete the user evaluation before the mean number of posts (34.7 for depression and 27.9 for eating disorders in \emph{test}). The maximum number of posts in \emph{test} indicates the total number of rounds for the evaluation stage. 

% (more details in \cite{marmol2023}). 

\begin{table}[!ht]
\caption{Details of the \emph{depression} and \emph{eating disorders} corpora. The number of users (total, positives, and negatives) and the number of posts per user (mean, minimum, and maximum) are reported.}
% \resizebox{\columnwidth}{!}{%
\begin{tabular}{|c|c|ccc|ccc|}
\hline
\multirow{2}{*}{} & \multirow{2}{*}{\textbf{ Corpus }} & \multicolumn{3}{c|}{\textbf{\#Users}} & \multicolumn{3}{c|}{\textbf{\#Posts per user}} \\ 
\cline{3-8} 
 & & \multicolumn{1}{c}{\textbf{ Total }} & \multicolumn{1}{c}{\textbf{ Pos }} & \textbf{ Neg } & \multicolumn{1}{c}{\textbf{ Mean }} & \multicolumn{1}{c}{\textbf{ Min }} & \textbf{ Max } \\ \hline
\multirow{3}{*}{\textit{\textbf{ Depression }}} & Train & \multicolumn{1}{c}{175} & \multicolumn{1}{c}{94} & 81 & \multicolumn{1}{c}{35.7} & \multicolumn{1}{c}{11} & 100 \\
 & Trial & \multicolumn{1}{c}{10} & \multicolumn{1}{c}{6} & 4 & \multicolumn{1}{c}{62.4} & \multicolumn{1}{c}{11} & 100 \\
 & Test & \multicolumn{1}{c}{149} & \multicolumn{1}{c}{68} & 81 & \multicolumn{1}{c}{34.7} & \multicolumn{1}{c}{11} & 100 \\ \hline
\multirow{3}{*}{\textit{\textbf{\begin{tabular}[c]{@{}c@{}}Eating \\ disorders \end{tabular}}}} & Train & \multicolumn{1}{c}{175}            & \multicolumn{1}{c}{74} & 101 & \multicolumn{1}{c}{33.9} & \multicolumn{1}{c}{11} & 50 \\
 & Trial & \multicolumn{1}{c}{10} & \multicolumn{1}{c}{5} & 5 & \multicolumn{1}{c}{38.9} & \multicolumn{1}{c}{18}           & 50           \\
 & Test & \multicolumn{1}{c}{150} & \multicolumn{1}{c}{64} & 86 & \multicolumn{1}{c}{27.9} & \multicolumn{1}{c}{11} & 50 \\ \hline
\end{tabular}%
% }
\label{fig:datasets}
\end{table}

We applied \emph{temporal fine-tuning} to the BETO model, a BERT variant trained on a large Spanish corpus \cite{CaneteCFP2020}, and ERDE30 as the training objective since the participants in MentalRiskES were ranked with this metric. \emph{Delays} and post windows were configured using 10 posts, and the rest of the hyperparameters were the following: learning\_rate=3e-5 (\emph{depression}) and 5e-5 (\emph{eating disorders}), and epoch=10, batch\_size=8, and optimizer=AdamW for both tasks. The best models were chosen by weighting the \emph{accuracy} and ERDE30 metrics. We used \emph{threshold} = 0.7 to detect positive users, and negative users were identified when they no longer had any posts. Besides, the \emph{minDelay} parameter was added to establish a minimum wait to start issuing alarms, testing \emph{minDelay}=10 and \emph{minDelay}=5.

We included in this study the baseline \emph{sliding\_window} model that was trained and validated considering the \emph{delay} scheme without including time in the input samples, along with the same hyperparameters and settings mentioned above. Table \ref{tab:results} shows the results obtained by our models. The three best results are included based on the ranking of the MentalRiskES Organizers according to ERDE30 among more than 25 proposals for both tasks\cite{marmol2023}. We also show the UNSL\#0 results (obtained by our laboratory in the challenge), which applied classic fine-tuning to the BETO model and a decision-making component based on the history of previous predictions \cite{thompson2023early}.

\begin{table}[!ht]
\caption{Results obtained considering the classification (Precision, Recall, and F1) and early classification (ERDE5, ERDE30, and F-latency) metrics for both tasks. The three best results are shown based on the ranking of MentalRiskES Organizers according to ERDE30, as well as the mean values among all results. 
% Values in bold depict the best performance for each metric.
Values in bold and underlined depict 1st and 2nd performance for each metric, respectively.
Values closer to 0 show better performances for the ERDE metric; for the rest of the metrics, values closer to 1 are preferred.}
\vspace{-5mm}
\begin{subtable}{0.95\textwidth} 
\centering  
\small
% \normalsize
\caption{Depression}
\label{subtab:depression_results}
\vspace{-1mm}
\begin{tabular}{lcccccc}
\hline
\textbf{Rank-Team\#Model} & \textbf{P}  & \textbf{R} & \textbf{F1} & \textbf{ERDE5$\downarrow$} & \textbf{ERDE30$\downarrow$} & \textbf{F-latency} \\
\hline
\noalign{\vspace{0.3mm}}
1-SINAI-SELA\#0 \cite{gonzalez2023sinai} & 0.78          & 0.74          & 0.72          & \underline{0.395}    & \textbf{0.140}  & \textbf{0.72}      \\
2-UNSL\#1 \cite{thompson2023early} & 0.79          & 0.76          & 0.73          & 0.567          & 0.148           & 0.61               \\
3-BaseLine-Deberta\#0 \cite{marmol2023} & 0.79          & 0.69          & 0.64          & \textbf{0.303} & 0.153           & \textbf{0.72}      \\
14-UNSL\#0                     & 0.75          & 0.74          & 0.73          & 0.551          & 0.188           & 0.59               \\
\textit{MentalRiskES2023-mean} & 0.73          & 0.66          & 0.62          & 0.383          & 0.232           & 0.60               \\ 
\noalign{\vspace{0.3mm}}
\hline
\noalign{\vspace{0.3mm}}
sliding\_window\_minDelay10    & 0.79          & 0.78          & 0.78          & 0.526          & 0.150           & 0.64               \\
sliding\_window\_minDelay5     & 0.80          & 0.76          & 0.75          & 0.473          & \textbf{0.140}  & \underline{0.71}         \\ 
\noalign{\vspace{0.3mm}}
\hline
\noalign{\vspace{0.3mm}}
temporal\_ft-minDelay10        & \textbf{0.83} & \textbf{0.83} & \textbf{0.83} & 0.486          & \underline{0.146}     & 0.66               \\
temporal\_ft-minDelay5         & \underline{0.81}    & \underline{0.81}    & \underline{0.81}    & 0.440          & 0.150           & \textbf{0.72}      \\ 
\noalign{\vspace{0.3mm}}
\hline
% \noalign{\vspace{-2mm}}
\end{tabular}  
% \caption{Depression}
% \label{subtab:depression_results}
% \vspace{-5mm}
\end{subtable}
\begin{subtable}{0.95\textwidth}
\centering
\small
% \normalsize
% \vspace{1mm}
\caption{Eating disorders}
\label{subtab:eating_results}
\vspace{-1mm}
\begin{tabular}{lcccccc}
\hline
\textbf{Rank-Team\#Model} & \textbf{P} & \textbf{R} & \textbf{F1} & \textbf{ERDE5$\downarrow$} & \textbf{ERDE30$\downarrow$} & \textbf{F-latency} \\ 
\hline
\noalign{\vspace{0.3mm}}
1-CIMAT-NLP-GTO\#0 \cite{echeverria2023cimat}          & \textbf{0.96} & \textbf{0.97} & \textbf{0.97} & \textbf{0.334} & \textbf{0.018}  & \textbf{0.86}      \\
2-UNSL\#1                   & \underline{0.91}    & \underline{0.92}    & \underline{0.91}    & 0.433          & \underline{0.045}     & \underline{0.78}         \\
3-CIMAT-NLP-GTO\#1          & 0.87          & 0.87          & 0.85          & 0.379          & 0.065           & 0.76               \\
11-UNSL\#0                  & 0.82          & 0.79          & 0.75          & 0.502          & 0.105           & 0.67               \\
\textit{MentalRiskES2023-mean} & 0.82          & 0.79          & 0.75          & 0.322          & 0.122           & 0.71               \\ 
\noalign{\vspace{0.3mm}}
\hline
\noalign{\vspace{0.3mm}}
sliding\_window\_minDelay10 & 0.85          & 0.85          & 0.85          & 0.454          & 0.108           & 0.61               \\
sliding\_window\_minDelay5  & 0.87          & 0.87          & 0.87          & 0.371          & 0.088           & 0.74               \\ 
\noalign{\vspace{0.3mm}}
\hline
\noalign{\vspace{0.3mm}}
temporal\_ft-minDelay10     & 0.90          & 0.90          & 0.90          & 0.448          & 0.069           & 0.65               \\
temporal\_ft-minDelay5      & \underline{0.91}    & 0.91          & \underline{0.91}    & \underline{0.365}    & 0.062           & 0.77               \\
\noalign{\vspace{0.3mm}}
\hline
% \noalign{\vspace{0.8mm}}
% \noalign{\vspace{-2mm}}
\end{tabular}
% \caption{Eating disorders}
% \label{subtab:eating_results}
\end{subtable}
\label{tab:results}
\end{table}

\subsection{Depression results}
The models obtained with \emph{temporal fine-tuning} achieved the second-best place by notably outperforming UNSL\#1, considering the overall performance among all metrics. 
In particular, \emph{temporal\_ft-minDelay10} obtained the second-best ERDE30 and the best classification performances, and \emph{temporal\_ft-minDelay5} achieved the best F-latency.
% and \emph{temporal\_ft-minDelay5} reached an F-latency closer to first place. 
It is highlighted that the \emph{temporal\_ft} models outperformed the respective \emph{sliding\_window} models in all metrics, with ERDE30 values very similar to each other. Additionally, our proposals remarkably outperformed UNSL\#0 and the mean performance among all teams across most metrics.

\subsection{Eating disorders results} achieved remarkable results
The \emph{temporal fine-tuning} models achieved remarkable results among the top positions, especially \emph{temporal\_ft-minDelay5}, which reached the third-best place, notably outperforming CIMAT-NLP-GTO\#1 in all metrics. The \emph{temporal\_ft-minDelay10} model was competitive with CIMAT-NLP-GTO\#1 on ERDE30, outperforming it in classification metrics. Besides, the \emph{temporal\_ft} models outperformed the respective \emph{sliding\_window} models in all metrics. Considering the mean among all teams, our results achieved better performance in Precision, Recall, F1, and ERDE30.

\subsection{Analysis of \emph{temporal fine-tuning} learning}
\begin{figure}[ht]
    \centering
    \includegraphics[width=0.95\textwidth]{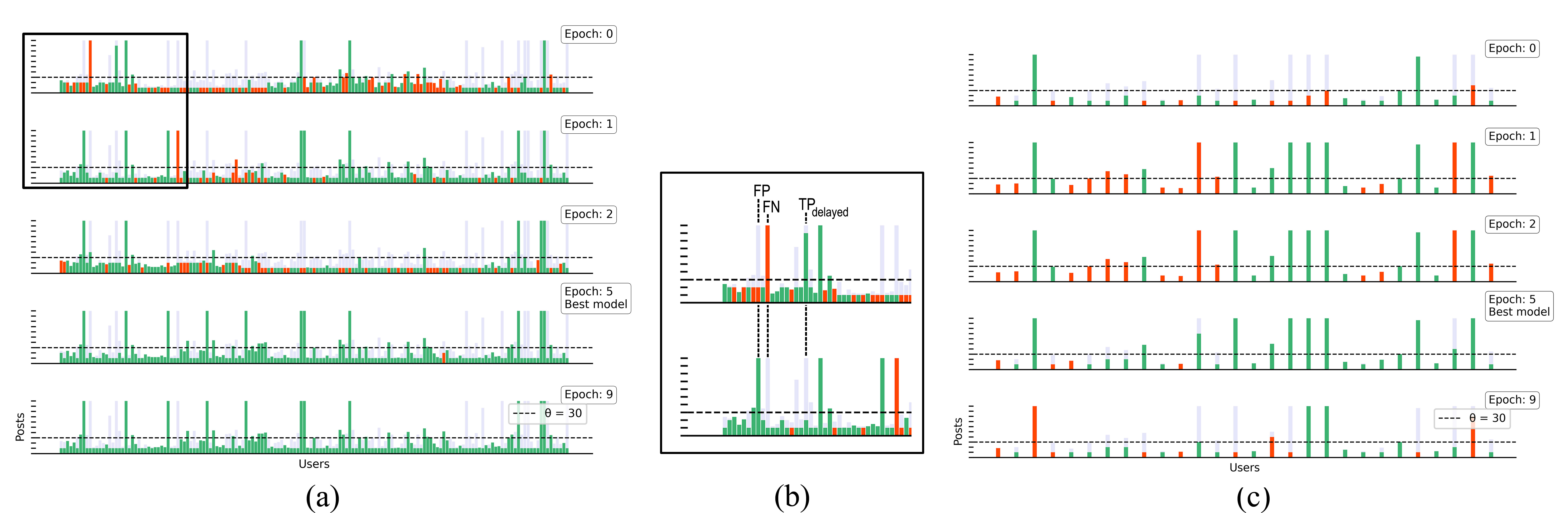} %\columnwidth .3\textwidth
    \caption{Model learning during \emph{temporal fine-tuning} for \emph{depression} task. (a) Training stage; (b) Some cases of (a) detected in epoch:0 and corrected in epoch:1; (c) Validation stage. On the x-axis, the vertical bars depict the model’s decisions, and their length shows the number of posts read and the instance of the final response. Green bar: correct decision; red bar: wrong decision; gray bar: unread posts during analysis. The y-axis shows \emph{delays} every 10 posts. The dashed horizontal line denotes the limit at $\theta$=30.}
    \label{fig:performance}
\end{figure}

We analyze the model behavior during learning using the depression task as an example. Fig. \ref{fig:performance}(a) shows that the model solves the early classification of users as the epochs progress. In the first epochs, the model tends to issue responses before $\theta$=30 due to the penalty. As epochs progress, final decisions are optimized by adjusting correctness and delay. Therefore, TPs tend to be obtained before $\theta$, while TNs when the user posts end, many of them in instances greater than $\theta$. Fig. \ref{fig:performance}(b) shows examples of FNs, FPs, and delayed TPs that were detected in epoch 0 and corrected in epoch 1. The best model of the validation stage was obtained at epoch 5 with optimal weighted performance (Fig. \ref{fig:performance}(c)). 
In this way, \emph{temporal fine-tuning} could learn when to end users' analysis during training, making it easier to adapt the model to an ERD environment without adding complex independent decision components.

\vspace{-0.4cm}
\subsection{\emph{Temporal fine-tuning} vs. mock-server}
\begin{figure}[ht]
\centering
\begin{subfigure}[b]{0.94\textwidth}
\centering
\includegraphics[width=0.94\textwidth]{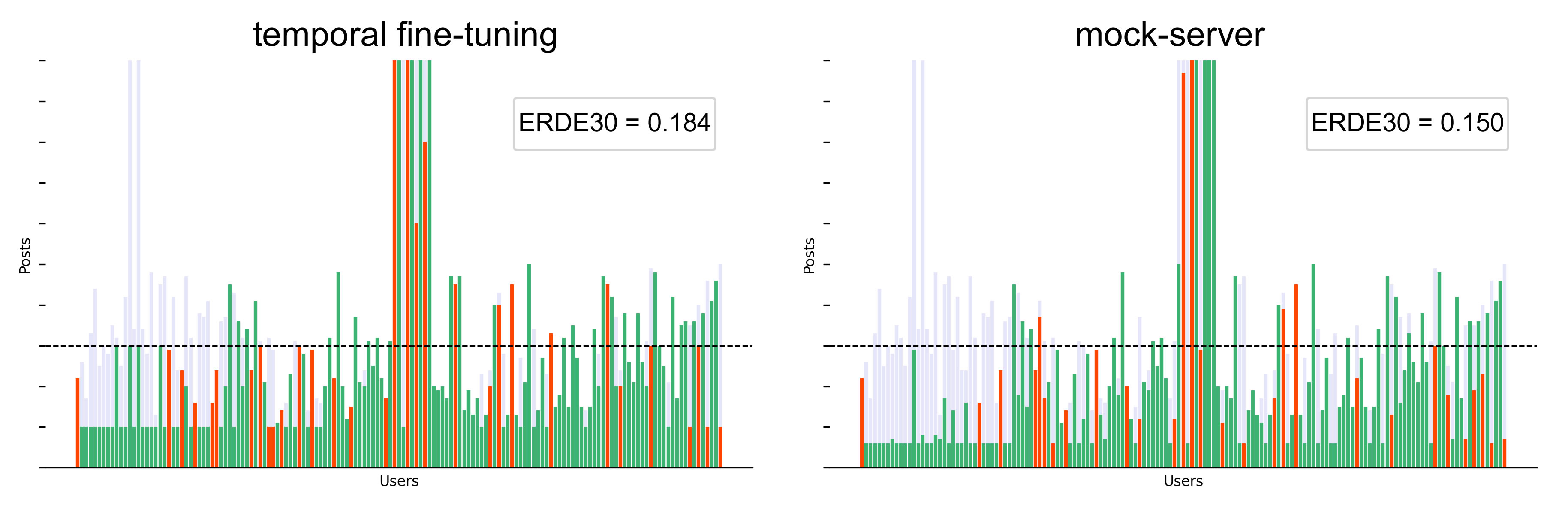}
\caption{Depression}
\label{fig:c}
\end{subfigure}
\begin{subfigure}[b]{0.95\textwidth}
\centering
\includegraphics[width=0.95\textwidth]{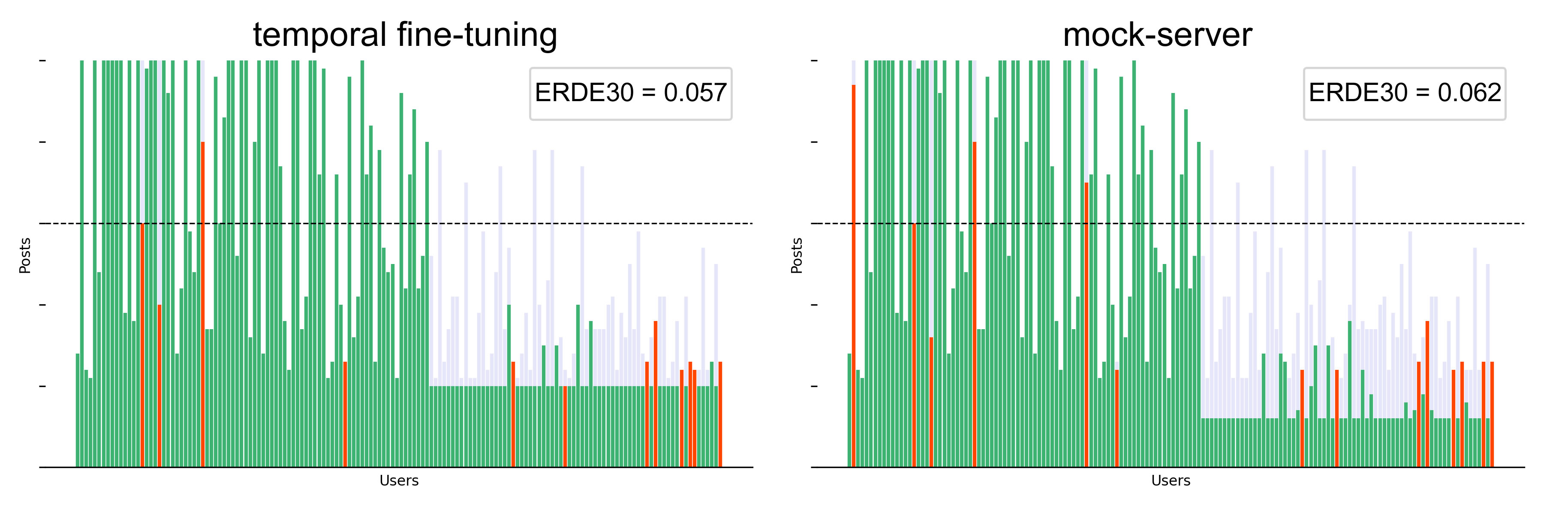}
\caption{Eating disorders}
\label{fig:b}
\end{subfigure}
\caption{Evaluation of the models using \emph{temporal fine-tuning} validation and mock-server for both tasks.}
\label{fig:temporal_vs_mockserver}
\end{figure}

Fig. \ref{fig:temporal_vs_mockserver} shows the models' behavior when evaluating the \emph{test} data using \emph{temporal fine-tuning} validation and mock-server. In the first, models can only decide every 10 posts, while in the mock-server, they decide post-by-post. Despite this, similar behaviors were obtained, demonstrating that our proposal design would allow us to know the expected performance in an ERD environment and appropriately select the best models. 

% \begin{figure}[!ht]
%     \centering
    
%     \begin{minipage}{\columnwidth}
%         \centering
%         \includegraphics[width=\columnwidth]{depression - temporal_mockserver.jpg}
%         \caption{(a) Depression} % Etiqueta sin número
%         % \label{fig:depression}
%     \end{minipage}
    
%     \vspace{1em} % Espacio vertical entre las figuras
    
%     \begin{minipage}{\columnwidth}
%         \centering
%         \includegraphics[width=\columnwidth]{eating_disorder - temporal_mockserver.jpg}
%         \caption*{(b) Eating disorders} % Etiqueta sin número
%         % \label{fig:eating_disorder}
%     \end{minipage}
    
%     \caption{Evaluation of the models using \emph{temporal fine-tuning} validation and mock-server for both tasks.}
%     \label{fig:temporal_vs_mockserver}
% \end{figure}

\subsection{Temporal representation}
\begin{figure}[!ht]
    \centering
    \includegraphics[width=.4\textwidth]{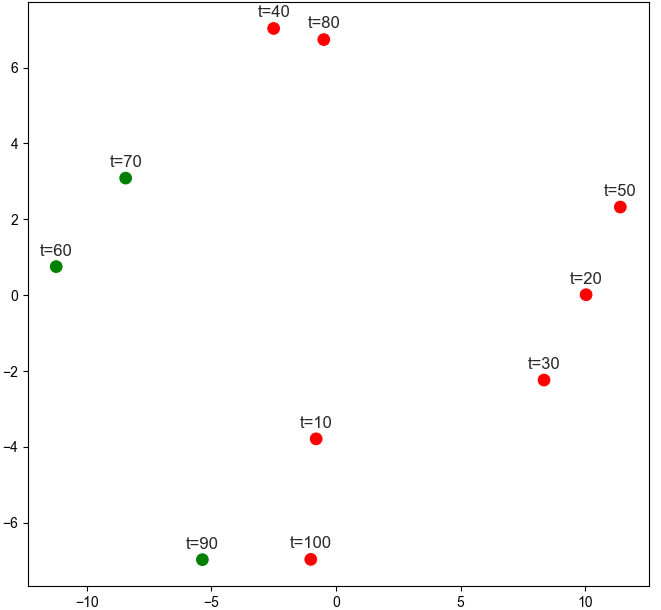} %\columnwidth .2\textwidth
    \caption{ Temporal representation of a sentence evaluated at times t=[10, 20, …, 100]. Green dot: positive decision; red dot: negative decision. Embeddings were extracted from the fine-tuned model ([CLS] token), followed by dimensionality reduction through Principal Component Analysis (PCA).}
    \label{fig:RepTemp}
\end{figure}

Due to the inclusion of time, the semantics of the words could change depending on the instance where they are analyzed. For example, for the sentence \emph{Hace mucho que no duermo bien...} (I haven't slept well for a long time…) evaluated at distinct times, different representations were obtained, directly influencing the classification result (Fig. \ref{fig:RepTemp}). A more in-depth analysis in the future will allow us to evaluate the impact of temporal representations and identify critical symptoms linked to diverse pathologies.

\subsection{Relevant themes in positive users}
A last analysis in our study consisted of obtaining the most relevant themes found in the positive users. Table \ref{tab:relevant_topics} shows the different themes found in positive users that were correctly detected. For each TP, we extracted the post window used by the models at the decision moment and used Google's Gemini model (\url{https://gemini.google.com/}) via API to capture the themes related to each task.

\begin{table}[!t]
\begin{center}
\caption{Themes relevant to depression and eating disorders tasks. For each positive user correctly detected, the post window used by the models at the decision moment was extracted. Then, they were analyzed by the Gemini model to capture the relevant themes.}
\label{tab:relevant_topics}
\begin{tabular}{ll}
\hline 
\noalign{\vspace{0.8mm}}
\textit{\textbf{Depression }}                                         & \begin{tabular}[c]{@{}l@{}}Loneliness, Depression, Suicidal ideation, \\ Hopelessness, Anxiety, Isolation, Sadness, \\ Low self-esteem, Loss and grief, Self-harm,\\  Eating disorders, Family conflicts, \\ Emotional dependence, Post-traumatic stress, \\ Sleep disorders, Loss of pleasure, \\ Medication dependency, Stress, \\ Lack of concentration, Fatigue,\\  Failed relationships.\end{tabular} \\ 
\noalign{\vspace{0.8mm}}
\hline 
\noalign{\vspace{0.8mm}}
\textit{\textbf{\begin{tabular}[c]{@{}l@{}}Eating \\ disorders \end{tabular}}} & \begin{tabular}[c]{@{}l@{}}Eating disorders, Body image, Bulimia, \\ Food restriction, Intermittent fasting, \\ Binge eating, Weight control, Advices, \\ Compensatory behaviors, Medical tests, \\ Physical exercise, Parental control, Self-harm,\\  Poor eating habits, Medication dependency, \\ Isolation, Relapses, Vomiting, Concerns, Diets.\end{tabular}\\ 
\noalign{\vspace{0.8mm}} 
\hline
\end{tabular}
\end{center}
\end{table}

\section{Conclusions and Future Work}
In this work, we proposed \emph{temporal fine-tuning}, a novel tuning method for pre-trained models based on transformers. We explicitly incorporated time during the learning process through a \emph{delay} scheme and a \emph{loss} function designed considering the ERDE$\theta$ metric. We found that our method optimizes the decisions, taking into account different contexts and time progress. It also allows us to evaluate training performance using temporal metrics and finally select the optimal models for an ERD problem.
We obtained remarkable results in the depression and eating disorders tasks for the Spanish language, being competitive with the best models of MentalRiskES 2023. In this way, by properly taking advantage of the power of transformers, it is possible to address ERD by combining precision and speed as a \emph{single objective}.
\setlength{\parskip}{0pt}

Although we have contributed to a less-explored language, it would be interesting to analyze our proposal in other languages, including English. Besides, new \emph{loss} functions could be designed considering other temporal metrics such as F-latency. On the other hand, it would be important to analyze the impact of the temporal representations obtained with our method, which would contribute to the interpretability of transformers. More research linked to the \emph{combined single-objective} approach is necessary since, in our view, it could be the most appropriate way to address ERD problems.

\begin{credits}
\subsubsection{\ackname} This work was developed at the Laboratorio de Investigación y Desarrollo en Inteligencia Computacional (LIDIC) and was
supported by a grant from Universidad Nacional de San Luis (UNSL), Argentina [PROICO 03-0620]. This work is part of the Doctoral thesis of Horacio Thompson.
\end{credits}


\bibliographystyle{splncs04}
\bibliography{paper} % Your .bib file without the extension





% LNCS Homepage, \url{http://www.springer.com/lncs}, last accessed 2023/10/25
% \end{thebibliography}

\end{document}